\documentclass[acmsmall]{acmart}
\usepackage{tabu}
\usepackage{subfigure}
\usepackage{color}
\AtBeginDocument{%
  }

\setcopyright{acmcopyright}
\copyrightyear{2023}
\acmYear{2023}
\acmDOI{XXXXXXX.XXXXXXX}

\acmJournal{JACM}
\acmVolume{37}
\acmNumber{4}
\acmArticle{111}
\acmMonth{8}

\begin{document}

\title{A Survey of Knowledge Enhanced Pre-trained Language Models}

\author{Jian Yang}
\affiliation{%
	\institution{Computer Science and Technology, Xidian University}
	\city{Xi'an}
	\country{China}
}
\email{seekerferry@gmail.com}


\author{Xinyu Hu}
\affiliation{%
	\institution{National Key Laboratory for Complex Systems Simulation}
	\city{Beijing}
	\country{China}
}
\email{husense@foxmail.com}


\author{Gang Xiao}
\authornote{Corresponding author.}
\affiliation{%
	\institution{National Key Laboratory for Complex Systems Simulation}
	\city{Beijing}
	\country{China}}
\email{searchware@qq.com}

\author{Yulong Shen}
\affiliation{%
	\institution{Computer Science and Technology, Xidian University}
	\city{Xi'an}
	\country{China}
}
\email{ylshen@mail.xidian.edu.cn}

\renewcommand{\shortauthors}{Yang et al.}

\begin{abstract}
Pre-trained language models learn informative word representations on a large-scale text corpus through self-supervised learning, which has achieved promising performance in fields of natural language processing (NLP) after fine-tuning. These models, however, suffer from poor robustness and lack of interpretability. We refer to pre-trained language models with knowledge injection as knowledge-enhanced pre-trained language models (KEPLMs). These models demonstrate deep understanding and logical reasoning and introduce interpretability. In this survey, we provide a comprehensive overview of KEPLMs in NLP. We first discuss the advancements in pre-trained language models and knowledge representation learning. Then we systematically categorize existing KEPLMs from three different perspectives. Finally, we outline some potential directions of KEPLMs for future research.
\end{abstract}

\ccsdesc[500]{Computing methodologies~Natural language processing}

\keywords{natural language processing, pre-trained language models, symbolic knowledge, knowledge enhanced pre-trained language models}

\received{20 February 2007}
\received[revised]{12 March 2009}
\received[accepted]{5 June 2009}

\maketitle
\section{Introduction}\label{sec1}
Current advances in deep learning \cite{1lecun2015deep, 2goodfellow2016deep, 3bengio2009learning}, pre-trained language models (PLMs) \cite{10radford2018improving,12devlin2019bert,15raffel2020exploring} in particular, have had an unprecedented impact both in the academic and industrial research communities. Deep learning leverages large-scale data through distributed representation and the hierarchical structure of neural networks. Based on deep learning, PLMs have undergone a qualitative leap forward, facilitating a wide range of downstream NLP applications. \par

Despite their achievements, there is still a long way to reach a new level of robust artificial intelligence. In the words of Marcus: intelligence can be counted on to apply what it knows to a wide range of problems systematically and reliably, synthesizing knowledge from a variety of sources such that it can reason flexibly and dynamically about the world, transferring what it learns in one context to another, in the way that we would expect of an ordinary adult \cite{new11marcus2020next}. Existing PLMs have the following limitations. First, they learn the semantics of the frequent words well but under-perform on rare words limited by long-tailed data distribution. Second, since PLMs are statistical models and learn implicit relations according to the co-recurrence signals, they are not adept at logical reasoning. Although PLMs can capture a wealth of linguistic \cite{new56jawahar2019does}, semantic \cite{new57yenicelik2020does}, syntactic \cite{new58hewitt2019structural} and even world knowledge \cite{131petroni2019language}, Cao et al. shows that the decent factual knowledge extraction performance of PLMs is largely due to the biased prompts \cite{re8cao-etal-2021-knowledgeable}. The experiments in \cite{re12talmor2020olmpics} also indicate the poor performance of PLMs in tasks requiring reasoning. Finally, ethical and societal concerns have been raised with PLMs outperforming humans in certain tasks. As observed, PLMs lack commonsense knowledge and generate illogical sentences \cite{new28lin2020commongen}. Thereby, interpretability and accountability of PLMs have become paramount for applying them generally. \par 

Combining neural networks with symbolic knowledge offers potential solutions to these problems. On the one hand, symbolic knowledge like knowledge graphs has high coverage of rare words, which addresses the lack of textual supervision \cite{86zhang2019ernie,60peters2019knowledge}. Furthermore, they provide comprehensive relational information \cite{new19lv2020graph,new30wang2017explicit} and/or explicit rules \cite{new26gangopadhyay2021semi} for models to enhance the reasoning of PLMs. On the other hand, symbolic knowledge improves interpretability of knowledge usage in downstream tasks \cite{new32amizadeh2020neuro}. In addition, it is practical to ingest knowledge\footnote{we use knowledge and symbolic knowledge interchangeably in this paper}  into a pre-trained checkpoint without training from scratch for a specific downstream application \cite{125liu2020k, new38marino2017more}. Therefore, researchers have explored integrating knowledge with PLMs to achieve more general artificial intelligence. \par

This paper provides a comprehensive overview of KEPLMs in NLP and discusses the pros and cons of each model in detail. The contributions of this survey can be summarized as follows:\par

\begin{itemize}
	\item \emph{Comprehensive review}. We offer a comprehensive review of pre-trained language models in NLP and delve into knowledge representation learning.\par
	\item \emph{New taxonomy}. We propose a taxonomy of KEPLMs, which categorizes existing KEPLMs from the granularity of knowledge, the method of knowledge injection, and degree of symbolic knowledge parameterization.\par
	\item \emph{Performance analysis}. We analyze the advantages and limitations for different categories of PLMs from the perspectives of the scope of application scenarios, effectiveness of knowledge injection, management of knowledge, and interpretability. 
	\item \emph{Future directions}. We discuss the challenges of existing KEPLMs and suggest some possible future research directions.\par
\end{itemize}\par
The rest of the survey is organized as follows. 
Section 2 outlines the progress of PLMs and knowledge representation learning. 
Section 3 introduces a classification principle and a corresponding comprehensive taxonomy. 
Following the categorization in Section 3, Section 4 introduces the working principle of each kind of KEPLMs and analyzes its pros and cons, 
and compares existing KEPLMs from different dimensions. 
Section 5 discusses the current challenges and suggests future directions.
{\color{black}All method abbreviations and their extensions proposed in the paper are listed in  Table \ref{table:abbr}.}
\begin{table*}[]
	\centering
	\normalsize
	\caption{{\color{black}Models abbreviations and corresponding extenstions}}
	\label{table:abbr} 
	\begin{tabular}{lllll}
	\cline{1-2}
	\multicolumn{1}{|l|}{{\color{black} Abbreviation }} & \multicolumn{1}{l|}{{\color{black}Full name }} &  &  &  \\ \cline{1-2}
	\multicolumn{1}{|l|}{{\color{black}NNLM }} & \multicolumn{1}{l|}{ {\color{black}Neural Network Language Model }} &  &  &  \\ \cline{1-2}
	\multicolumn{1}{|l|}{{\color{black}CBOW }} & \multicolumn{1}{l|}{ {\color{black}Continuous Bag-of-Words }} &  &  &  \\ \cline{1-2}
	\multicolumn{1}{|l|}{{\color{black}SG }} & \multicolumn{1}{l|}{ {\color{black}Skip-Gram }} &  &  &  \\ \cline{1-2}
	\multicolumn{1}{|l|}{{\color{black}GloVe }}  & \multicolumn{1}{l|}{ {\color{black}Global Vectors}} &  &  &  \\ \cline{1-2}
	\multicolumn{1}{|l|}{{\color{black}LSTM }} & \multicolumn{1}{l|}{ {\color{black}Long Short-Term Memory}} &  &  &  \\ \cline{1-2}
	\multicolumn{1}{|l|}{{\color{black}ELMo }} & \multicolumn{1}{l|}{ {\color{black}Embeddings from Language Models}} &  &  &  \\ \cline{1-2}
	\multicolumn{1}{|l|}{{\color{black}ULMFiT }} & \multicolumn{1}{l|}{ {\color{black}Universal Language Model Fine-tuning}} &  &  &  \\ \cline{1-2}
	\multicolumn{1}{|l|}{{\color{black}GPT }} & \multicolumn{1}{l|}{ {\color{black}Generative Pre-trained Transformer }} &  &  &  \\ \cline{1-2}
	\multicolumn{1}{|l|}{{\color{black}BERT }} & \multicolumn{1}{l|}{ {\color{black}Bidirectional Encoder Representations from Transformers}} &  &  &  \\ \cline{1-2}
	\multicolumn{1}{|l|}{{\color{black}RoBERTa }} & \multicolumn{1}{l|}{ {\color{black}A Robustly Optimized BERT Pretraining Approach}} &  &  &  \\ \cline{1-2}
	\multicolumn{1}{|l|}{{\color{black}T5 }} & \multicolumn{1}{l|}{ {\color{black}Text-to-text Transfer Transformer}} &  &  &  \\ \cline{1-2}
	\multicolumn{1}{|l|}{{\color{black}TreansE }} & \multicolumn{1}{l|}{ {\color{black}Translations in the Embedding Space}} &  &  &  \\ \cline{1-2}
	\multicolumn{1}{|l|}{{\color{black}TransH	}} & \multicolumn{1}{l|}{ {\color{black}Translation on Hyperplanes}} &  &  &  \\ \cline{1-2}
	\multicolumn{1}{|l|}{{\color{black}TransR	}} & \multicolumn{1}{l|}{ {\color{black}Translation in the Relation Space}} &  &  &  \\ \cline{1-2}
	\multicolumn{1}{|l|}{{\color{black}HolE}} & \multicolumn{1}{l|}{ {\color{black}Holographic Embeddings}} &  &  &  \\ \cline{1-2}
	\multicolumn{1}{|l|}{{\color{black}SME}} & \multicolumn{1}{l|}{ {\color{black}Structured Mixture of Experts}} &  &  &  \\ \cline{1-2}
	\multicolumn{1}{|l|}{{\color{black}NTN}} & \multicolumn{1}{l|}{ {\color{black}Neural Tensor Network}} &  &  &  \\ \cline{1-2}
	\multicolumn{1}{|l|}{{\color{black}R-GCN}} & \multicolumn{1}{l|}{ {\color{black}Relational Graph Convolutional Network}} &  &  &  \\ \cline{1-2}
	\multicolumn{1}{|l|}{{\color{black}SACN  }} & \multicolumn{1}{l|}{ {\color{black}Self-Attention based Convolutional Networkr}} &  &  &  \\ \cline{1-2}
	\multicolumn{1}{|l|}{{\color{black}ConvE  }} & \multicolumn{1}{l|}{ {\color{black}Convolutional 2D Knowledge Graph Embeddings }} &  &  &  \\ \cline{1-2}
	\multicolumn{1}{|l|}{{\color{black}CompGCN}} & \multicolumn{1}{l|}{ {\color{black}Compositional Graph Convolutional Network  }} &  &  &  \\ \cline{1-2}
	\multicolumn{1}{|l|}{{\color{black}SentiLARE		}} & \multicolumn{1}{l|}{ {\color{black}Sentiment-aware Language Representation}} &  &  &  \\ \cline{1-2}
	\multicolumn{1}{|l|}{{\color{black}ERNIE(Baidu)}} & \multicolumn{1}{l|}{ {\color{black}Enhanced Representation through Knowledge Integration}} &  &  &  \\ \cline{1-2}
	\multicolumn{1}{|l|}{{\color{black}ZEN}} & \multicolumn{1}{l|}{ {\color{black}Chinese Text Encoder Enhanced by Representing N-grams}} &  &  &  \\ \cline{1-2}
	\multicolumn{1}{|l|}{{\color{black}LUKE	}} & \multicolumn{1}{l|}{ {\color{black}Language Understanding with Knowledge-based Embeddings}} &  &  &  \\ \cline{1-2}
	\multicolumn{1}{|l|}{{\color{black}ERNIE(Tingshua) }} & \multicolumn{1}{l|}{ {\color{black}Enhanced Language Representation with Informative Entities}} &  &  &  \\ \cline{1-2}
	\multicolumn{1}{|l|}{{\color{black}KnowBERT	}} & \multicolumn{1}{l|}{ {\color{black}Knowledge Enhanced BERT }} &  &  &  \\ \cline{1-2}
	\multicolumn{1}{|l|}{{\color{black}BRET-MK	 }} & \multicolumn{1}{l|}{ {\color{black}BERT-based Language Model with Medical Knowledge }} &  &  &  \\ \cline{1-2}
	\multicolumn{1}{|l|}{{\color{black}K-BERT	 }} & \multicolumn{1}{l|}{ {\color{black}Knowledge-enabled Bidirectional Encoder Representation from Transformers }} &  &  &  \\ \cline{1-2}
	\multicolumn{1}{|l|}{{\color{black}CoLAKE		}} & \multicolumn{1}{l|}{ {\color{black}Contextualized Language and Knowledge Embedding}} &  &  &  \\ \cline{1-2}
	\multicolumn{1}{|l|}{{\color{black}COMET	}} & \multicolumn{1}{l|}{ {\color{black}Common Metadata for Educational Resources}} &  &  &  \\ \cline{1-2}
	\multicolumn{1}{|l|}{{\color{black}T5+SSM	}} & \multicolumn{1}{l|}{ {\color{black}Text-to-text Transfer Transformer with Salient Span Masking}} &  &  &  \\ \cline{1-2}
	\multicolumn{1}{|l|}{{\color{black}WKLM	}} & \multicolumn{1}{l|}{ {\color{black}Weakly Supervised Knowledge-pretrained Language Model}} &  &  &  \\ \cline{1-2}
	\multicolumn{1}{|l|}{{\color{black}LIBERT	 }} & \multicolumn{1}{l|}{ {\color{black}Lexically Informed BERT}} &  &  &  \\ \cline{1-2}
	\multicolumn{1}{|l|}{{\color{black}GLM	 }} & \multicolumn{1}{l|}{ {\color{black}Graph-guided Masked Language Model}} &  &  &  \\ \cline{1-2}
	\multicolumn{1}{|l|}{{\color{black}KEPLER	 }} & \multicolumn{1}{l|}{ {\color{black}Knowledge Embedding and Pre-trained Language Representation}} &  &  &  \\ \cline{1-2}
	\multicolumn{1}{|l|}{{\color{black}K-ADAPTER		}} & \multicolumn{1}{l|}{ {\color{black}Infusing Knowledge into Pre-trained Models with Adapters}} &  &  &  \\ \cline{1-2}
	\multicolumn{1}{|l|}{{\color{black}ERICA	 }} & \multicolumn{1}{l|}{ {\color{black}Entity and Relation Understanding via Contrastive Learning}} &  &  &  \\ \cline{1-2}
	\multicolumn{1}{|l|}{{\color{black}LIMIT-BERT	 }} & \multicolumn{1}{l|}{ {\color{black}Linguistics Informed Multi-task BERT}} &  &  &  \\ \cline{1-2}
	\multicolumn{1}{|l|}{{\color{black}SKEP	 }} & \multicolumn{1}{l|}{ {\color{black}Sentiment Knowledge Enhanced Pre-training}} &  &  &  \\ \cline{1-2}
	\multicolumn{1}{|l|}{{\color{black}KT-NET		 }} & \multicolumn{1}{l|}{ {\color{black}Knowledge and Text Fusion Net}} &  &  &  \\ \cline{1-2}
	\multicolumn{1}{|l|}{{\color{black}KGLM	 }} & \multicolumn{1}{l|}{ {\color{black}Knowledge Graph Language Model}} &  &  &  \\ \cline{1-2}
	\end{tabular}
	\end{table*}

\section{Related Work}
\label{sec:rw}

\subsection{Pre-trained Language Models}
\label{subsec:hb}
We refer to the model that can extract high-level features from a large unsupervised text corpus and obtain effective representations, which can then be applied to downstream tasks after fine-tuning, as a pre-trained language model. %
The effectiveness of the pre-trained language model largely depends on the  representation learning of the model's encoder. %
Representation learning refers to learning representations of the data that make it easier to extract useful information when building classifiers or other predictors \cite{23bengio2013representation}. %
There are two mainstream paradigms within the community of representation learning: probabilistic graphical models and neural networks. %
Probabilistic graph models learn feature representation by modeling the posterior distribution of potential variables in sample data, including directed graph model and undirected graph model. %
Most neural network models utilize an autoencoder, which is composed of both an encoder and a decoder. %
The encoder is responsible for feature extraction, while the decoder reconstructs the input by applying a regularized reconstruction objective.\par
Neural network-based models are preferable to probabilistic graphical models for the following reasons. %
Firstly, neural networks can express more possible features with distributed vectors instead of sparse vectors. %
Secondly, considering the existing data is mainly the result of the interaction between multiple latent factors,  distributed vectors can represent different impact factors by designing a specific network structure. %
Finally, the underlying neural layers of deep neural networks transform concrete features learned from data into abstract features in upper layers and remain stable even with local changes in the input data, enhancing the robustness of representation for generalization in various downstream tasks.\par%
Following autoencoder-based neural models, pre-trained language models design specific neural networks to encode input data while using pre-trained tasks to decode learned representations. %
After fine-tuning, pre-trained language models can easily be adapted to all kinds of NLP tasks. %
We categorize models into token-based and context-based pre-trained language models based on whether they capture sequence-level semantics.\par

\subsubsection{Token-based Pre-trained Language Models}
Originating from the NNLM \cite{4bengio2003neural} that was proposed by Bengio in 2003, the training process inherently generates distributed representations of words as a by-product. %
According to the hypothesis that words with similar context have similar semantics, Mikolov et al. propose two shallow architectures: Continuous Bag-of-Words (CBOW) and Skip-Gram (SG) to capture latent syntactic and semantic similarities between words \cite{5mikolov2013efficient,6mikolov2013distributed}. %
\textcolor{black}{Both models are constructed using  a three-layer neural network, 
consisting of input, hidden, and output layers.}
Besides, GloVe \cite{7pennington2014glove} computes word-word cooccurrence statistics from a large corpus as a supervised signal, and FastText \cite{8bojanowski2017enriching} trains the model with text classification data. %
With the emergence of all the above token-based pre-trained language models, word embeddings have been commonly used as text representations in NLP tasks.  %
Although these models are simple and effective, they are only suited to attain fixed representations rather than capturing polysemy. %
That is also why we call this type of model static pre-trained language models. %
\subsubsection{Context-based Pre-trained Language Models}
To address the problem of polysemy, pre-trained language models need to distinguish the semantics of words and dynamically generate word embeddings in different contexts. %
Given a text $x_1$,$x_2$,$\cdots$,$x_T$ where each token $x_t$ is a word or sub-word, the contextual representation of $x_t$ depends on the whole text. %
\begin{align}
	\label{1}
	[h_1,h_2,\cdots,h_T]=f_{enc}(x_1,x_2,\cdots,x_T),
\end{align}
where $f_{enc}(\cdot)$ is a neural encoder and $h_t$ is a contextual embedding.\par %
Taking LSTM \cite{24hochreiter1997long} as a neural encoder, the ELMo extracts context-dependent representations from a bidirectional language model, which has shown to bring large improvement on a range of NLP tasks \cite{9peters2018deep}. However, ELMo is usually used as a feature extractor to produce initial embeddings for the specific model of downstream tasks, which means the rest parameters of the main model have to train from scratch. \par
Around the same time, the introduction of ULMFiT  offered valuable insights into multi-stage transfer and the fine-tuning of models \cite{25howard2018universal}. %
Besides, Transformer has achieved surprising  success on machine translation and proven to be more effective than LSTM in dealing with long-range text dependencies \cite{26vaswani2017attention}. %
In this background, OpenAI proposes GPT \cite{10radford2018improving}  that adopts the modified Transformer's decoder as a language model to learn universal representations transferable to a broad range of downstream tasks, which outperforms task-specific architectures in 9 of 12 NLP tasks. %
GPT-2 and GPT-3 \cite{11radford2019language,12-brownlanguage} mainly follow the architecture and train on larger and more diverse datasets to learn from varied domains. %
However, limited by a unidirectional encoder, the GPT series can only attend its left context resulting in sub-optimal for learning sentence-level semantics. %
To overcome this deficiency, BERT \cite{12devlin2019bert}  adopts a masked language modelling (MLM) objective where some of the tokens of a sequence are  masked randomly, and the goal is to predict these tokens considering the corrupted sentence. %
Inspired by Skip-Thoughts \cite{27kiros2015skip}, BERT also employs the next sentence prediction (NSP) task to learn the semantic connection between sentences, which obtains new start-of-art results on eleven NLP tasks and even becomes the basis of subsequent models. %
Based on BERT, RoBERTa \cite{14liu2019roberta} design a few improved training recipes, including training longer with bigger batches over more data, modifying objectives, training over long sequences, and dynamically changing the masking pattern, which enhances significantly performance of BERT. %
To overcome the discrepancy between pre-training and fine-tuning of BERT, XLNet \cite{13yang2019xlnet} proposes a new autoregressive method based on permutation language modeling to capture contextual information without introducing any new symbols.\par %
Unlike all these above pre-trained language models that aim at natural understanding or generation tasks, T5 \cite{15raffel2020exploring} adopts an encoder-decoder framework to unify natural understanding and generation tasks by converting the data into the text-to-text format. 
\subsubsection{Analysis}
In the development progress of pre-trained language models, the critical step is the change from only attaining the fixed word representations to train the whole neural architecture. %
The emergency of GPT introduces the paradigm of pretraining and fine-tuning , enabling pre-trained language models to be transferable to downstream tasks without the need for training from scratch. The subsequent works focus on expanding the range of text that pre-trained language models can attend, improving the efficiency of models, and generalizing more downstream scenarios. For more details of models, we refer readers to \cite{28liu2020survey,29qiu2020pre}.

\subsection{Knowledge Representation}
In this section, we first introduce the definition of knowledge and then the conventional methods of knowledge representation, and comprehensive knowledge representation learning based on them.
\subsubsection{Knowledge}
Knowledge is a familiarity, awareness, or understanding of someone or something, such as facts (descriptive knowledge), skills (procedural knowledge), or objects. David et al. divided knowledge into four categories, namely factual knowledge, conceptual knowledge, procedural knowledge, and metacognitive knowledge \cite{30krathwohl2002revision}. Factual knowledge refers to the knowledge of terminology and specific details and elements to describe objective things. Conceptual knowledge is the interrelationships among the fundamental elements within a larger structure that enables them to function together, such as principles, generalizations, and theories. Procedural knowledge is about the knowledge that guides action, including methods of inquiry and criteria for using skills, algorithms, techniques, and methods. Metacognitive knowledge emphasizes self-initiative and is the knowledge of cognition in general as well as awareness.\par 
\subsubsection{Methods of Knowledge Representation}
Davis et al. put forward the definition of knowledge representation in 1993, arguing that the notion can best be understood in terms of five distinct roles  \cite{31davis1993knowledge}. First, a knowledge representation is most fundamentally a surrogate, a substitute for the thing itself, and we can reason about the world through thinking without practice. Second, it is a set of ontological commitments about how to think about the world. Third, it is a fragmentary theory of intelligent reasoning. Fourth, it is a medium for pragmatically efficient computation, which  supports recommended inferences through the effective knowledge organization. Fifth, it is the medium of human expression used to express cognition of the world.\par
Conventional knowledge representation methods include first-order predicate logic, frame representation \cite{32minsky2019framework}, script representation \cite{33tomkins1978script}, semantic network representation \cite{34quillan1966semantic}, and ontology representation. The basic grammatical elements of first-order predicate logic are symbols representing objects, relations, and functions, among which the objects refer to the individual or category of things, the relationships refer to the mapping between things, and the functions require the object in each domain to have a mapping value as a special form of a predicate. %
Although this method can guarantee the consistency of knowledge representation and the correctness of inference results, it is difficult to represent procedural knowledge. \par%
A semantic network is a conceptual network represented by a directed graph where nodes represent concepts and edges represent semantic relations between concepts, which can also be transformed into triplets. %
It can describe knowledge in a unified and straightforward way that is beneficial for computer storage and retrieval. However, it can only represent conceptual knowledge but not dynamic knowledge such as procedure knowledge.  \par%
The frame representation organizes knowledge hierarchically, where each entity is denoted by a frame that contains multiple slots to store attributes and their corresponding values. It avoids duplicate definitions of the frame by inheriting one's property. %
Due to the diversity and complexity of the real world, many actual situations and frameworks differ greatly  introducing errors or conflicts in the framework design process, which causes the lack of generality except its inability to represent procedural knowledge. \par 
Script representation represents the basic behaviour of things through a series of atomic actions, which describes the occurrence of things in a definite temporal or causal order and is used for  dynamic knowledge. Although it can represent procedural knowledge to a certain extent, it is not appropriate for conceptual or factual knowledge. \par%
Originally, the term ``ontology'' comes from philosophy, where it is used to describe the existence of beings in the world. For the sake of obtaining models with reasoning capabilities, researchers adopt the term ontology to describe what can be computationally represented of the world in a program. %
CYC \cite{36lenat1993building} is a knowledge base constructed following ontology specifications, aiming to organize human commonsense knowledge. Since ontologies can represent unanimously recognized static domain knowledge, it is also used in information retrieval and NLP. WordNet \cite{37miller1995wordnet} is created based on word ontologies. In addition to static knowledge modelling, task-specific ontologies are also designed to add reasoning capabilities based on static knowledge. \par%
In order to promote semantic understanding, Tim et al. propose the Semantic Web concept in 2001 to build a massive distributed database that links data through semantics instead of strings  \cite{38berners2001semantic}. %
To make data understandable for computers, W3C proposes the Resource Description Framework (RDF) that uses the semantic network representation to express semantics in the form of triples  \cite{160miller1998introduction}. %
This form can be easily implemented by applying graph algorithms of probability graphs and graph theories. %
Besides, Web Ontology Language (OWL) is designed to enable computers reasoning ability, which describes categories, attributes and instances of things complying with ontology representation. \par%
In the implementation of engineering solutions, the knowledge graph (KG) is the knowledge base represented as a network with entities as nodes and relations as edges. %
Specifically, KGs obtain knowledge and corresponding descriptions from the network by semantic web technology and is organized as triplets. %
Since the procedural knowledge is hard to manage and its certainty is weak,  most of the existing KGs only contain conceptual and factual knowledge without procedural knowledge. %

\subsubsection{Knowledge Representation Learning}
Knowledge representation learning (KRL) delegated by deep learning focuses on representation learning of entities and relations in the knowledge base, which effectively measures semantic correlations of entities and relations and alleviates sparsity issues. More importantly, symbolic knowledge can be much easier to integrate with the neural network based models after knowledge representation learning.\par

\paragraph{Translational Distance Models} With distance-based scoring functions, this type of model measures the plausibility of a fact as the distance between the two entities after a translation carried out by the relation. %
Inspired by linguistic regularities presented in \cite{161mikolov2013linguistic}, TransE represents entities and relations in $d$-dimension vector space so that the embedded entities $h$ and $t$ can be connected by translation vector $r$, i.e., $h+r\approx t$ when $(h,r,t)$ holds \cite{45bordes2013translating}. %
To tackle this problem of insufficiency of a single space for both entities and relations, TransH and TransR allow an entity to have distinct representations when involved in different relations  \cite{46wang2014knowledge,47ji2015knowledge}. TransH introduces relational hyperplanes assuming that entities and relations share the same semantic space, while TransR exploits separated space for relations to consider different attributes of entities. %
TransD argues that entities serve as different types even with the same relations and construct dynamic mapping matrices by considering the interactions between entities and relations \cite{48lin2015learning}. %
Owing to heterogeneity and imbalance of entities and relations, TranSparse simplifies TransR
by enforcing sparseness on the projection matrix  \cite{49ji2016knowledge}. %

\paragraph{Semantic Matching Models} 
Semantic matching models measure plausibility of facts by matching latent semantics of entities and relations with similarity-based scoring functions. %
RESCAL associates each entity and relation with a vector and matrix, repectively \cite{40nickel2011three}. The score of a fact $(h,r,t)$ is defined by a bilinear function. %
\textcolor{black}{The bilinear function introduces matrices to represent relations between entities, enabling the modeling of local structure centered around relations within knowledge graphs.}
To decrease the computing complexity, DistMult simplifies RESCAL by restricting relation to diagonal matrices  \cite{41yang2014embedding}. %
Combining the expressive power of RESCAL with the efficiency and simplicity of DistMult, HolE composes the entity representations with the circular correlation operation, and the compositional vector is then matched with the relation representation to score the triplet  \cite{42nickel2016holographic}. %
Unlike models above, SME conducts semantic matching between entities and relations using neural network architectures  \cite{43bordes2014semantic}. %
NTN combines projected entities  with a relational tensor and predicts scores after a relational linear output layer  \cite{44socher2013reasoning}. %

\paragraph{Graph Neural Network Models} The above models embed entities and relations by only facts stored as a collection of triplets, while graph neural network based models take account of the whole structure of the graph. %
Graph convolutional network (GCN) was first proposed in \cite{165bruna2014spectral} and has become an effective tool to create node embeddings  which aggregates local neighborhood information of the graphs \cite{162kipf2016semi,163kipf2016variational,164hamilton2017inductive,167velivckovic2017graph}. %
As the extension of graph convolutional networks, R-GCN is developed to deal with the highly multi-relational data characteristic of realistic knowledge bases  \cite{50schlichtkrull2018modeling}. %
SACN employs an end-to-end network learning framework where the encoder leverages graph node structure and attributes  \cite{51shang2019end}, and the decoder simplifies ConvE \cite{166dettmers2018convolutional} and keeps the translational property of TransE. %
Following the same framework of SACN, Nathani et al. propose an attention-based feature embedding that captures both entity and relation features in the encoder \cite{53nathani2019learning}. %
Vashishth et al. believe that the combination of relations and nodes should be considered comprehensively during the message transmission  \cite{56vashishth2019composition}. Therefore they propose CompGCN that leverages various entity-relation composition operations from knowledge graph embedding techniques and scales with the number of relations to embed both nodes and relations jointly. %

\section{Classification of Knowledge Enhanced Pre-trained Language Models}
To gain insight into how to attain effective KEPLMs, we performed a systematic classification of existing KEPLMs based on broad a comprehensive literature review. The objectives are to offer guidelines for users and researchers by comparing their pros and cons. \par

\subsection{The Principle of Classification}
Symbolic knowledge offers a wealth of insights in the form of entity descriptions, KGs, and rules for PLMs. This knowledge enhances entity features, strengthens inter-entity associations, and provides guidance for the inference processes in PLMs. PLMs require different granularities of knowledge to address various downstream tasks. For example, entity-level knowledge, which includes aspects like sentiment polarity and supersenses, enables PLMs to proficiently tackle tasks like entity recognition and word disambiguation. In contrast, structured knowledge aids models in structural prediction and question answering tasks. However, merely possessing knowledge is insufficient for PLMs. An efficient method of knowledge injection is also essential. Given that different forms of knowledge have distinct representation methods, models must consider their data structures to design appropriate knowledge injection techniques. The method of knowledge injection can greatly influence the effectiveness of knowledge integration. For unstructured knowledge, which shares a data structure with inputs, models can use the same encoder to process both simultaneously, eliminating concerns about heterogeneity. However, structured knowledge necessitates additional techniques for vector space learning. Furthermore, different degrees of knowledge parameterization greatly influence the interpretability and reliability of PLMs.
Much effort has been devoted to examining  the knowledge encoded in PLMs by different probing ways \cite{172rogers2020primer}. 
Researchers have found that token representations of PLMs can capture syntactic and semantic knowledge by probing classifiers \cite{new62hewitt2019structural,new63jawahar2019does}. The quantitative analysis in question answering tasks demonstrates that PLMs can encode structured commonsense knowledge \cite{173cui2021commonsense}. Clark et al. explore the functions of self-attention heads and report that they attend to words significantly in certain syntactic positions \cite{re11clark2019does}. Despite these achievements,  no study has delved into the interpretability of how knowledge is utilized in downstream tasks, especially for tasks requiring intensive knowledge. 
\textcolor{black}{Besides, due to the lack of a modeling for formal logic and a rigorous reasoning process, PLMs achieve subpar performance in question answering tasks that involve multi-hop reasoning and logical reasoning tasks over knowledge bases. When confronted with multi-hop question answering tasks, it is challenging for PLMs to dynamically determine which part of a question to analyze at each hop and predict the corresponding relations, often leading to incorrect or incomplete answers. Similarly, without a structured methodology to manage geometric operations like projection, intersection, and union, PLMs may struggle to accurately execute inductive logical reasoning \cite{DBLP:conf/acl/WangWHFSZH23}.}
Thus, we categorize existing KEPLMs from the following three dimensions: the granularity of knowledge, the method of knowledge injection, and the degree of symbolic knowledge parameterization. \par%

\subsection{A Taxonomy of Knowledge Enhanced Pre-trained Language Models}
This section gives a concrete taxonomy according to dimensions discussed above. %
\subsubsection{The Granularity of Knowledge}
KEPLMs integrate different granularity of knowledge for use in scenarios that require information at different detail levels. In general, sentiment analysis mainly relies on word features and thus requires more information about individual entities. In contrast, the text generation task relies on commonsense knowledge, and the question answering task relies on rules and KGs for inference.We classify KEPLMs based on the granularity of integrated knowledge into: entity-fused, syntax-tree-fused, KG-fused, and rule-fused categories. \par 

\paragraph{Entity-fused KEPLMs}
As a basic semantic unit, an entity exists in words, phrases, and literals. 
\textcolor{black}{KEPLMs typically project entity features into embeddings, which are combined with word embeddings via pooling operations. The linguistic and lexical knowledge associated with entities enables PLMs to proficiently address fine-grained tasks such as named entity recognition \cite{re7li2020enhancing}, sentiment analysis \cite{58ke-etal-2020-sentilare} and word disambiguation tasks.}
\paragraph{Text-fused KEPLMs}
\textcolor{black}{Texts often contain detailed descriptions of entities, relationships, and events. KEPLMs use texts as external references, extracting critical information that primarily benefits question answering tasks. These models encode both questions and texts into embeddings and extract the most pertinent details from vast external references through semantic similarity computations. Based on a knowledge-augmented input that concatenates the relevant information with the question, they subsequently forecast the most probable answers.} \par 

\paragraph{Syntax-tree-fused KEPLMs} 
Syntactic knowledge presents the important constituents of a sentence. 
\textcolor{black}{For syntax trees, KEPLMs either transform them into semantic representations using suitable representation learning methods and combine them with word representations  \cite{new4sachan2021syntax}, or utilize them to select critical constituents from input sequences for masking. This form of knowledge proves beneficial for structure-aware tasks, including syntactic parsing \cite{new2zhou2020limit}, semantic role labeling, and relation extraction.}\par

\paragraph{KG-fused KEPLMs} 
With the advance of the technique of information extraction, plenty of general and domain-specific KGs have emerged. KGs provide a structured way to represent rich information in the form of entities and relations between them. 
\textcolor{black}{One approach involves representation learning for KGs and subsequently merging their representations with aligned word representations using an infusion module \cite{60peters2019knowledge,86zhang2019ernie}. Another approach transforms KGs and sequences into a unified data structure, encoding them with the same PLM. 
With rich and detailed entity knowledge, KG-fused KEPLMs are versatile, covering a wide range of language understanding and generation tasks.} \par

\paragraph{Rule-fused KEPLMs}  
Rules can be characterized as either formal guidelines or rigorous logical formulations. Their primary advantage lies in the interpretability and accountability offered by strict mathematical structures and a transparent inference process. 
\textcolor{black}{KEPLMs employ symbolic reasoning post-module, drawing from their prediction outcomes, which is essential for reasoning tasks emphasizing transparency and accountability.}

\subsubsection{The Method of Knowledge Injection}
The method of knowledge injection plays an important role in the effectiveness and efficiency of integration between PLMs and knowledge, as well as the management and storage of knowledge. In fact, it dictates the type of knowledge that can be integrated and its form. To gain insight into how knowledge is injected, we classify models into feature-fused,  embedding-combined, knowledge-supervised, data-structure-unified, retrieval-based and rule-guided KEPLMs.  \par   

\paragraph{Feature-fused KEPLMs}
This type of model obtains features such as the sentiment polarity, the supersense,  and the span of entities from a specific knowledge base. Feature-fused KEPLMs usually take it into account by projecting into embedding with a trainable matrix and learn its meaning by pre-training task \cite{174sun2019ernie, 175sun2020ernie, 58ke-etal-2020-sentilare}.

\paragraph{Embedding-combined KEPLMs}
To fill the gap between symbolic knowledge and neural networks, embedding-combined KEPLMs transform symbolic knowledge into embedding with a representation learning algorithm in advance that sharply influences the performance of models. Then the tokens in text and entities will be aligned to combine the corresponding embedding of them by attention mechanism or other weighting operations \cite{86zhang2019ernie,60peters2019knowledge}. 
However, there will be heterogeneous semantic space because of different representation learning algorithms for different forms of knowledge. To solve this problem, some KEPLMs generate the initial embedding of nodes with their contexts \cite{128wang2021kepler,new19lv2020graph}. \par 

\paragraph{Data-structure-unified KEPLMs}
Due to structural incompatibility, some works adopt different representation learning algorithms for knowledge injected and original training data of PLMs. However, it causes the heterogeneous semantic space and increases the difficulty of fusion of them. To integrate both of them smoothly, data-structure-unified KEPLMs convert the relational triplets of the KG to sequences, thus the same encoder is used for learning embeddings \cite{22sun2020colake,125liu2020k,99he2020integrating}. However, the construction of a unified data structure relies on heuristic realization and the structural information of the KG is discarded. 

\paragraph{Knowledge-supervised KEPLMs}
To avoid extra training costs and engineering design, knowledge-supervised KEPLMs choose the entities that meet a specific relation and/or relational triplets as training data \cite{58ke-etal-2020-sentilare,128wang2021kepler}. As we discuss above, PLMs are statistical models and learn relations between entities by co-recurrence signals. To overcomes the issue, KEPLMs concatenate relational triplets and/or entity with the input sequence without sacrificing efficiency \cite{new12yamada2020luke,new13DBLP:conf/acl/QinLT00JHS020}.

\paragraph{Retrieval-based KEPLMs}
Instead of injecting knowledge, retrieval-based KEPLMs can update perception by consulting external knowledge. They usually retrieve desired information from knowledge sources by computing the relevance between input text and knowledge \cite{new9joshi2020contextualized,142guu2020realm,new59DBLP:conf/nips/LewisPPPKGKLYR020}. One of the strengths lies in the initiative of choosing relevant information, which avoids the effect of redundant and ambiguous knowledge that cannot match the input text. Since they do not preserve knowledge within models, they are limited in their application and mainly apply for question answering. 

\paragraph{Rule-guided KEPLMs}
Most of the KEPLMs store knowledge and language information within the parameters of the model. However, it is not intuitive to observe how the knowledge is leveraged in downstream tasks. A straightforward approach is to maintain the original form of symbolic knowledge, as rule-guided KEPLMs do. This type of model consists of perception and reasoning systems where the former is made up with PLMs and the latter is achieved with rules \cite{new60han2021ptr}. A major advantage of such models is that they guarantee the reliability of results using rigorous mathematical formulations and provide interpretability through an explicit reasoning process. %

\subsubsection{The Degree of Knowledge Parameterization}
Knowledge can be harnessed by PLMs in different parameterized degrees. 
\textcolor{black}{ Fully parameterized knowledge enables the effective integration of knowledge with PLMs and end-to-end training, facilitating efficient training and direct utilization of models without requiring coordination with additional modules.}
To bridge the symbolic knowledge and neural networks, the former is projected into a dense, low-dimensional semantic space and presented by distributed vectors through knowledge representation learning \cite{21ji2021survey}. 
Current algorithms mainly focus on representation learning toward KGs. Variants of GNN are employed to capture the structure of KGs. 
\textcolor{black}{However, this approach also presents challenges concerning knowledge management and interpretability in usage. The introduction of new information or modifications often necessitates computationally expensive retraining or vector representation adjustments. Conversely, maintaining symbolic knowledge forms enables efficient updates and interpretability, meeting downstream task requirements for knowledge timeliness and transparency of knowledge usage. However, models that maintain symbolic knowledge forms face challenges in bridging the gap between symbolic knowledge and neural computations, and they also confront optimization issues. This is especially true for hybrid systems that aim to combine rule-based logic with PLMs.}
 According to the degree of knowledge parameterization, we divide models into fully parameterized, partially parameterized, and knowledge form unchanged KEPLMs. %

\paragraph{Fully parameterized KEPLMs}
With the rapid development of GNN, various symbolic knowledge, especially for KGs, can be encoded effectively. 
\textcolor{black}{Improved techniques in the aggregation and combination stages of GNNs, including attention mechanisms, hierarchical aggregations, and gating mechanisms, have resulted in more comprehensive and selective  encoding of nodes and relations within KGs.}
Not only for the semantics of entities, this type of model also captures structure information by virtue of the superior method of KRL to support reasoning. By storing knowledge as parameters, models can be knowledge-aware and adaptable to a broad range of scenarios.%

\paragraph{Partial parameterized KEPLMs}
Owing to the limitation of GNN for modeling multi-step relations of KGs, partial parameterized KEPLMs encode only a portion of the knowledge while keeping the rest of it unchanged. For instance, some works \cite{new30wang2017explicit,136verga2021adaptable} encode the entities of KGs but maintain the structural information in the symbolic form. The representation learning of entities is responsible for integration with PLMs, while the structure information of KGs is responsible for retrieving the associated entities. This is particularly efficient for obtaining as many related entities as possible to support decisions considering the massive amount of relational triplets in KGs.

\paragraph{Knowledge form unchanged KEPLMs}  
In addition to outstanding performance, researchers have recognized the need for offering a better understanding of the underlying principles of KEPLMs. %
Rule-based representation provides a mapping mechanism between symbols and PLMs. By integrating symbol reasoning system into the learning pipeline, knowledge form unchanged KEPLMs reconcile the advantages of effective perception of PLMs and reasoning and interpretability of symbolic representation \cite{new60han2021ptr}.

The taxonomy and the corresponding KEPLMs introduced in the paper are shown in Fig. \ref{Taxonomy}. %

\section{Overview of Knowledge Enhanced pre-trained language models}
In this section, we give a detailed account of the KEPLMs we found in our literature survey. We focus on methods of knowledge injection and, therefore, organize our presentation according to this dimension. This is based on the premise that the method of knowledge injection, being the central factor, dictates the types of knowledge PLMs can integrate and the form in which this knowledge appears. Following this idea, we introduce existing KEPLMs from the perspective of application scenarios, knowledge injection efficiency, knowledge management, and interpretability in knowledge usage. \par 
To visualize the association between the methods of knowledge injection, the type of knowledge, and the degree of knowledge parameterization, we have drawn a schematic diagram for each type of KEPLM. The line thickness of the graph represents the quantity of each type. \par %

\subsection{Feature-fused KEPLMs}
Feature-fused KEPLMs focus on entity feature utilization. %
\textcolor{black}{In general, entity features are either projected into embeddings 
	and merged with word embeddings through pooling operations or they are used to determine masked subtokens for pre-training tasks. %
	The correlation between knowledge injection methods, various types of knowledge, and the form of knowledge is presented in Fig. \ref{section1}. %
\begin{figure}[H]
	\centerline{\includegraphics[width=5.6 in]{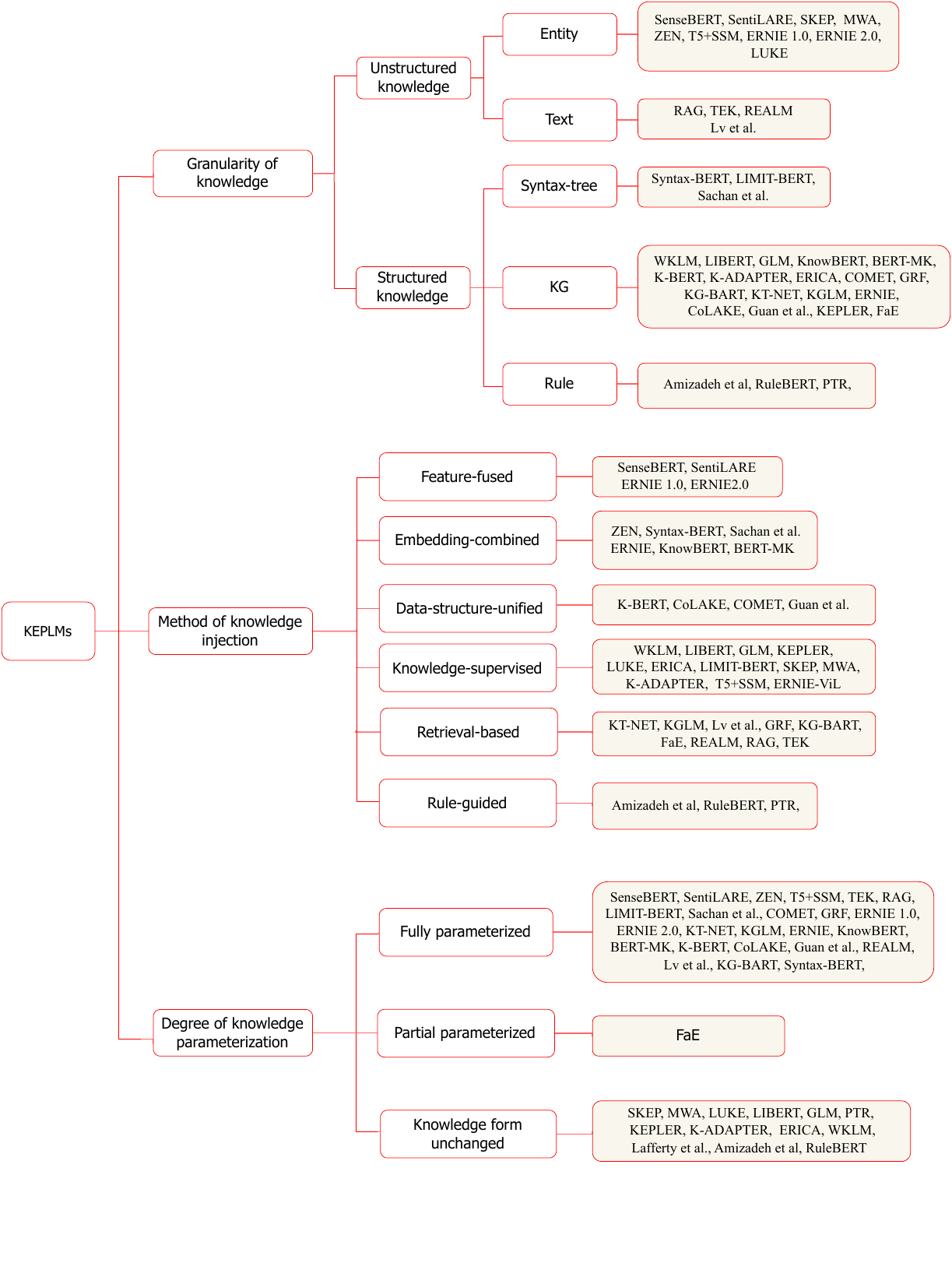}}
	\caption{Taxonomy of KEPLMs with Representative Examples.\label{Taxonomy}} 
\end{figure}
\begin{figure}[t] 
	\includegraphics[width=3.5in]{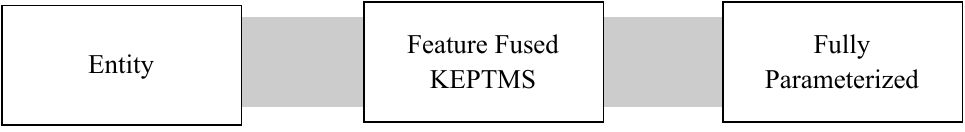} 
	\caption{Feature-fused KEPLMs integrate entity knowledge by fully parameterizing it.}
	\label{section1}
\end{figure} 
\begin{figure}[t]
	\centering
	\begin{minipage}[t]{0.5\linewidth}
		\centering
		\includegraphics[width=2.3in]{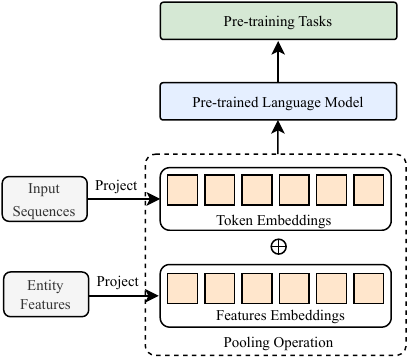}
		\caption{Infusing knowledge and PLMs via  \\ projecting knowledge-related features into \\ embeddings.} 
		\label{survey1}
	\end{minipage}%
	\begin{minipage}[t]{0.5\linewidth}
		\centering
		\includegraphics[width=2.5in]{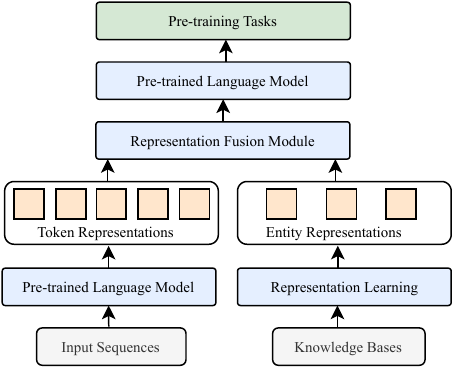}
		\caption{Infusing knowledge and PLMs via representation learning at advance and specific knowledge fusion module.}
		\label{survey2}
	\end{minipage}%
\end{figure}
Models like SenseBERT \cite{57levine2020sensebert} and SentiLARE \cite{58ke-etal-2020-sentilare} project sentiment polarity and supersenses 
into an embedding space, combining them with the original embedding at the input layer, 
and then predict entity features with infused representations in pre-training tasks. The knowledge integration process is shown in Fig. \ref{survey1}.} 
\textcolor{black}{ Conversely, models such as ERNIE 1.0 \cite{174sun2019ernie} leverage entity boundary information 
to determine the masked span, 
thereby orienting the PLMs toward entity prediction during the pre-training phase. 
The integration provides more detailed and comprehensive entity knowledge, 
which is especially significant when PLMs struggle to extract such information 
solely from context. 
As a result, this integration empowers PLMs to excel in intricate entity-centered tasks, 
including named entity recognition, sentiment analysis and word disambiguation tasks.}\par  
\textcolor{black}{SenseBERT introduces a word sense prediction task that involves ascertaining the accurate senses of a word with representations 
fusing word representations and word-sense features to incorporate word-sense information. 
Specifically, SenseBERT utilizes the WordNet word-sense inventory to annotate word supersenses 
and projects them into embeddings. 
These embeddings are integrated with word embeddings by addition operation 
after dimension transformation. 
Subsequently, SenseBERT predicts the senses of words with the integrated representations 
during the pre-training.}
\textcolor{black}{By explicitly exposing the model to lexical semantics, SenseBERT can avoid the ambiguity in word forms. Therefore, SenseBERT considerably outperforms a vanilla BERT on a Supersense Disambiguation task and the Word in Context task \cite{59pilehvar2019wic} that requires word supersense awareness.}
Although BERT has been proven successful in simple sentiment classification, directly applying it to fined-grained sentiment analysis shows less significant improvement \cite{61sun2019utilizing}. %
Therefore, to better solve the above issue, SentiLARE integrates word-level linguistic knowledge into pre-trained models for sentiment analysis \cite{58ke-etal-2020-sentilare}. %
Taking RoBERTa as the backbone model, SentiLARE first acquires the part-of-speech tags and computes word sentiment polarity from SentiWordNet by a context-aware attention mechanism. %
\textcolor{black}{The context-aware attention mechanism determines the attention weight of senses by considering the sense rank and the similarity between the context and the gloss of each sense to capture multiple senses of words.}
Then two pre-training tasks are utilized to capture the relationship between sentence-level language representation and word-level linguistic knowledge. %
\textcolor{black}{Sentence-level language representation encapsulates the broader semantic and contextual comprehension of a text or a sentence, reflecting its global sentiment or meaning. }\textcolor{black}{Conversely, word-level linguistic knowledge is centered around individual words, capturing specific attributes such as part-of-speech tags and sentiment polarity.}
\textcolor{black}{To assess its efficacy, SentiLARE is benchmarked against both task-specific pre-trained models and models without pre-training across aspect-level and sentence-level sentiment analysis tasks. The improved outcomes, as reflected in the F1 score and accuracy metrics, underscore the effectiveness of injecting sentiment knowledge.}\par
Limited by the word segmentation methods, %
\textcolor{black}{PLMs break down entities into subtokens that often have different semantics than the original entities, posing a challenge for accurate comprehension of entity meanings. For instance, in the case of the song ``Blowing in the Wind'', PLMs might interpret it as a description of an actual wind blowing rather than the song by Bob Dylan. This can impact their performance in tasks such as entity recognition and subsequent relationship classification.}
To overcome the issue, ERNIE 1.0 employs entity and phrase masking strategies to tell the span of semantic units and learns the embeddings of them by the context \cite{174sun2019ernie}. Its improved version, ERNIE 2.0 introduces different prediction or classification pre-training tasks to capture lexical, syntactic and semantic information simultaneously  \cite{175sun2020ernie}. 
Notably, \textcolor{black}{ERNIE 2.0 adopts a continual pre-training framework that initializes the parameters of the model trained with previous pre-training tasks when introducing a new task to achieve incremental learning.}
Utilizing the method, ERNIE 2.0 suggests that multitask learning techniques could be a viable strategy for integrating multiple knowledge sources into PLMs. 
Catastrophic forgetting is a common phenomenon when diverse knowledge is learned by PLMs. To this end, the multi-task learning technique is prioritized to integrate multiple knowledge into pre-trained language models. PLMs can benefit from a regularization effect to alleviate overfitting to a specific task, thus making the learned representations universal across tasks.

\subsection{Embedding-combined KEPLMs}
Although feature-fused KEPLMs can learn the rich semantics of entities, it is challenging to perform reasoning with entities alone. 
\textcolor{black}{Reasoning intrinsically requires comprehending the relations, dynamics, and contexts in which entities operate. Merely understanding the meaning of entities does not provide insights into their interactions with other entities or their role within a broader context. For instance, Yasunaga et al.  emphasize the necessity of fusing relation information from KGs to encourage joint reasoning over text and KGs \cite{yasunaga2022deep}. Consequently, capturing relational data between entities is imperative for PLMs' reasoning capabilities.}\par
To capture KG's knowledge, embedding-combined KEPLMs encode them using KRL in advance and infuse corresponding embeddings by specific fusion module, as shown in Fig. \ref{survey2}. They leverage a broader range of knowledge such as entities, syntax-trees, and KGs, and store knowledge in the parameter form. After equipping knowledge, embedding-combined KEPLMs are applied to general natural language understanding, and question answering. 
\begin{figure}[htbp] 
	\centering 
	\includegraphics[width=3.5in]{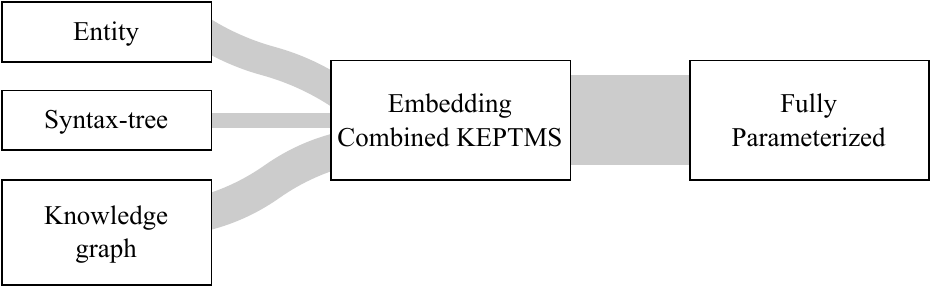} 
	\caption{All embedding-combined KEPLMs transform knowledge into neural network parameters. While the majority harness knowledge from entities and knowledge graphs, a minority utilize knowledge derived from syntactic trees.} 
	\label{section2}
\end{figure} 
\textcolor{black}{
	The correlation between knowledge injection methods, various types of knowledge, and the extent of knowledge parameterization is shown in Fig. \ref{section2}.}\par %
The span masking strategy is popular for injecting boundary features of entities. However, it can only infuse a single entity for each aligned token embedding and results in a mismatch between pre-training and fine-tuning. %
To avoid these issues and further utilize the semantics within the spans, Li et al. propose a multi-source word aligned attention (MWA) to integrate explicit word information with pre-trained character embeddings  \cite{re7li2020enhancing}. 
\textcolor{black}{The MWA utilizes information about the subtoken composition of words to determine the scope of attention for each subtoken, 
leading to a reallocation of attention scores. 
Specifically, an initial self-attention operation is performed on the input character representations, 
yielding a matrix of character-level attention scores. 
Subsequently, the input sequence is segmented into non-overlapping spans using word segmentation tools, 
and a mixed pooling strategy \cite{new6yu2014mixed} is employed to calculate inner-word attention using the character-level attention score matrix. 
Finally, the enhanced character representations are generated with a modified attention mechanism that substitutes the original attention score matrix with the new one.
By recomputing the attention vector of one character from the perspective of the whole word, 
the attention bias caused by character ambiguity can be eliminated effectively.}
Unlike the previous model, ZEN
learns entity representations with an external encoder instead of reassigning attention scores for
entities \cite{re6diao2020zen}. 
\textcolor{black}{Compared to reassigning attention scores,
 employing an external encoder not only treats entities as basic semantic units 
 but also facilitates mutual attention among multiple salient n-grams within a sentence with self-attention mechanisms.}
To learn more granular text, ZEN considers
different combinations of characters during pre-training by attending n-gram representations.
Given a sequence of Chinese characters, the model extracts n-grams and records their positions
with an n-gram matching matrix. Then all n-grams are encoded by a Transformer encoder and then combined
with associated characters. Compared to the models that adopt masking strategies, ZEN and MWA can incorporate nested entities and thus significantly improve entity
integration’s generality while affording little training cost.
Different from the above models, LUKE employs an extra vocabulary to record embeddings of entities \cite{new12yamada2020luke}. It treats words and entities as independent tokens and computes representations for all tokens using the Transformer. Concretely, it uses a large amount of entity-annotated corpus obtained from Wikipedia. Considering the vast cost and computational efficiency, the authors decompose entity embeddings into two smaller matrices. 
\textcolor{black}{Besides, an entity-aware self-attention mechanism has been introduced to distinguish between words and entities. This is achieved by calculating attention scores for every potential token pair using distinct query matrices.}
Since entities are treated as complete semantic units, LUKE is able to directly model the relationships between entities and achieves strong empirical performance in knowledge-driven NLP tasks. \par 

Beyond entities, syntax-tree can also be utilized to enhance PLMs. %
Syntactic biases are beneficial for various natural language understanding tasks that require structured output spaces, such as semantic role labeling and coreference resolution. The Syntax-BERT model efficiently integrates syntactic knowledge into pre-trained Transformers by employing sparse mask matrices, which capture different syntactic relationships of the input. This integration is achieved through a syntax-aware self-attention mechanism \cite{new1bai2021syntax}. In contrast to this method, Sachan et al. utilize a graph neural network to encode the dependency structure of the input sentence \cite{new4sachan2021syntax}. Given that BERT employs subwords as input units rather than traditional linguistic tokens, their model modifies the original dependency tree. This is done by introducing additional edges, specifically by linking the first subword of a token to its subsequent subwords.\par

\begin{figure}[htbp] 
	\centering 
	\includegraphics[width=3.5in,height=2.5in]{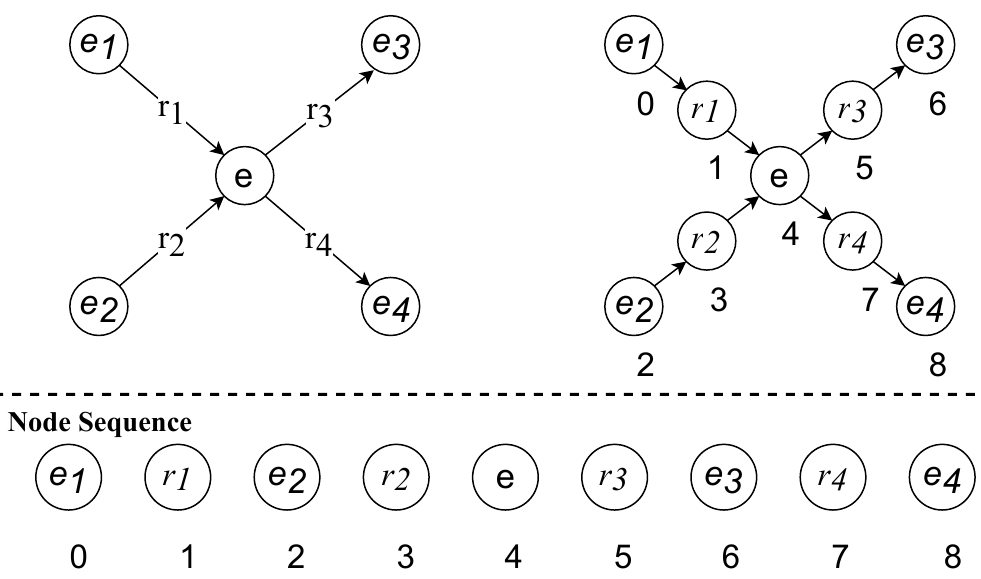} 
	\caption{ The Conversion of the Subgraph \cite{99he2020integrating}.} 
	\label{BERT-MK} 
\end{figure}

As the most common knowledge, KGs provide a comprehensive and rich information of entities and relations. Different representation learning algorithms are proposed to attain its embeddings. ERNIE encodes the entities and relations using  TransE  \cite{45bordes2013translating} and integrates entity representations and token embeddings based on the alignments by self-attention mechanism \cite{86zhang2019ernie}. %
\textcolor{black}{In a similar vein, KnowBERT recontextualizes word representations by attending to entity embeddings learned in advance and word embeddings with an attention mechanism to capture factual knowledge of all entities \cite{60peters2019knowledge}. Rather than leveraging existing alignment data, KnowBERT employs an auxiliary entity linker trained with minimal supervision data to discern a broader set of entity mentions of sentences. The auxiliary entity linker introduces a mention-span self-attention to compute mention representations, enabling KnowBert to integrate global information into each linking decision. Subsequently, the model incorporates entity information into the mention-span representations through a weighted summation of their representations. Finally, the model recontextualizes word representations using an adapted transformer layer that substitutes the multi-headed with a multi-headed attention between aligned words and knowledge-augmented entity-span vectors. By integrating the factual knowledge of entities, KnowBERT demonstrates improved performance in knowledge-driven tasks, notably relationship extraction and entity typing.}
After integrating relational triplets of the KG into BERT, both of the model demonstrate improved ability to recall facts in knowledge-driven tasks like relationship extraction, entity typing.  %
However, they treat triplets as a independent training unit during KRL procedure, ignoring the informative neighbors of entities. %
BERT-MK \cite{99he2020integrating} captures richer semantic of triplets from KGs by utilizing contextualized information of the nodes. %
The subgraphs of entities are extracted from the KG and transformed into a sequence, which is shown in Fig. \ref{BERT-MK}. %
Considering mutual influence of entities and relations, the relations are regarded as graph nodes as well. %
The sequence of nodes is subsequently input into the Transformer and a knowledge integration framework analogous to ERNIE is employed to integrate entity contextual information.
However, not all knowledge plays a positive role in the KEPLMs. Redundant and ambiguous knowledge in KGs will be injected when KEPLMs encode sub-graphs independently of textual context. To this end, CokeBERT dynamically selects contextual knowledge and embed knowledge context according to textual context \cite{new23su2021cokebert}. \par%
\begin{figure}[htbp] 
	\centering 
	\includegraphics[width=3.5in]{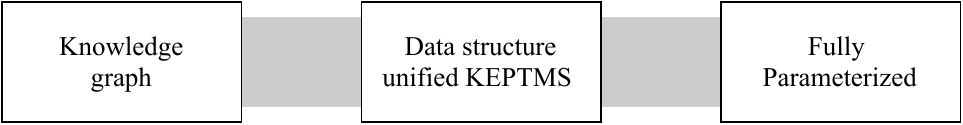} 
	\caption{Date strucutre unified KEPLMs integrate knowledge graphs by fully parameterizing it.}
	\label{section3}
\end{figure} 
Overall, most entity combined KEPLMs typically undergo two preliminary steps for knowledge integration: knowledge representation learning and alignment. However, there are some errors in alignments for tokens and entities. 
\textcolor{black}{Since entities often possess multiple aliases and are composed of several tokens, ambiguities or morphological variations of entities can lead to incorrect mapping of entities to tokens that have entirely different semantics.}
Therefore, it is critical to enhance the robustness of KEPLM to capture knowledge under misalignment situations. For instance, ERNIE is asked to predict correct entities based on wrong alignments introduced deliberately \cite{86zhang2019ernie}. Conventional KRL methods typically process triples in isolation, overlooking the intricate details that may be implicit in the local context surrounding a triple. In contrast, GNNs are more adept at encoding structural knowledge. The knowledge injection technique discussed in this section is adaptable across various knowledge granularities. Additionally, embedding-combined KEPLMs incorporate knowledge within model parameters, making them versatile for various application contexts. However, they also present challenges. One significant drawback is the increased computational demand required to represent and integrate diverse knowledge forms. This knowledge injection approach also complicates the assurance that the model acquires specific knowledge. It also hinders our ability to explicitly update or remove knowledge. When essential information updates, embedding-combined KEPLMs require retraining to preserve knowledge accuracy, leading to suboptimal knowledge management. Adapter modules offer a potential solution to ease the challenges of knowledge updating. They store different knowledge types within individual adapters, requiring minimal trainable parameters. This design facilitates the addition of new knowledge without the need to alter existing ones. Furthermore, the original PLMs' parameters remain unchanged, promoting significant parameter sharing. \par 

\subsection{Data-structure-unified KEPLMs}
\begin{figure}[htbp]
	\centering
	\begin{minipage}[t]{0.5\linewidth}
		\centering
		\includegraphics[width=2in]{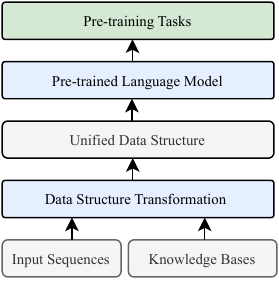}
		\caption{Infusing knowledge and PLMs via unifying \\ the structure of input.} 
		\label{survey3}
	\end{minipage}%
	\begin{minipage}[t]{0.5\linewidth}
		\centering
		\includegraphics[width=2.9in]{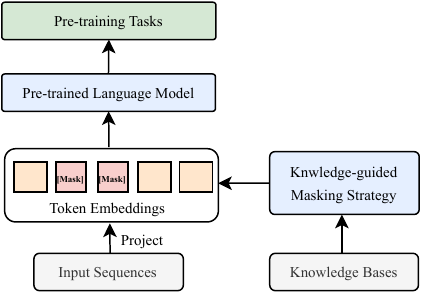}
		\caption{Infusing knowledge and PLMs via emphasizing knowledge of input.} 
		\label{survey4}
	\end{minipage}%
\end{figure}

For accommodating different structure of text and KGs, data-structure-unified KEPLMs transform sequences and knowledge into a unified structure and encode embeddings with the same encoder to avoid heterogeneous vector space. 
\textcolor{black}{The correlation between knowledge injection methods, various types of knowledge, and the extent of knowledge parameterization is presented in Fig. \ref{section3}.} The type of model enhances PLMs mainly via KGs and captures knowledge by storing them within model parameters. The knowledge integration process is shown in Fig. \ref{survey3}.\par
\begin{figure} [htbp]
	\centering 
	\includegraphics[width=3.5in]{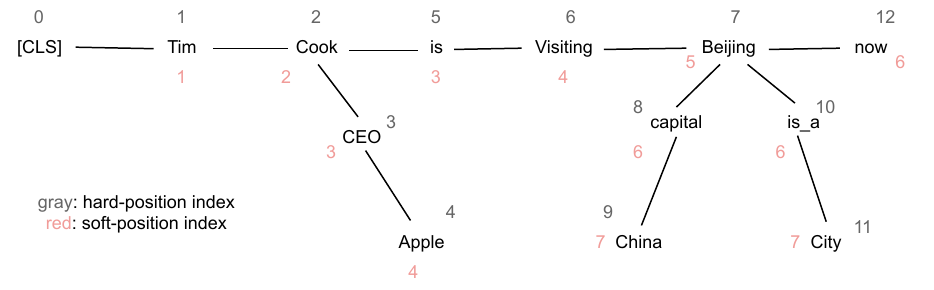} 
	\caption{The Structure of the Sentence Tree \cite{125liu2020k}.} 
	\label{k-bert} 
\end{figure}

K-BERT integrates knowledge by connecting sequences with relevant triples, forming a knowledge-rich sentence tree \cite{125liu2020k}. %
Specifically, all the entity mentions involved in the sentence are selected out to query  corresponding triples in KGs, and then K-BERT stitches the triples to corresponding positions to generate a sentence tree shown in Fig. \ref{k-bert}. %
Without considering the inconsistency of structure, K-BERT infuses the associated information of entities by fine-tuning on downstream tasks and achieves 1-2\% $F$1 gains in specific domain tasks.
Notably, when fine-tuned with CN-DBpedia \cite{126xu2017cn}, K-BERT performs better than when fine-tuned with HowNet in question answering and named entity recognition, which demonstrates the importance of an appropriate KG for different scenarios. %
Although K-BERT infuses triplets and sequences by employing a unified data structure, it treats relational triplets as independent units and ignores the associations between them. %
\textcolor{black}{Contextual triplets can provide critical entity knowledge for PLMs, particularly in instances where these models struggle to infer entity semantics only from context.}
To this end, CoLAKE constructs a word-knowledge graph and integrates contextual triplets  through a pre-training task \cite{22sun2020colake}. %
\textcolor{black}{Utilizing a word-knowledge graph that bridges sentence sequences and triplets, 
CoLAKE's pre-training task predicts masked words, 
entities and relations simultaneously within the given context to integrate contextual triplets. 
In detail, CoLAKE constructs the word-knowledge graph by transforming the input sequence 
into a connected word graph and concatenating this graph and sub-graphs of aligned entities. 
Following BERT's masking strategy, CoLAKE masks word, entity, and relation nodes and predict them given their context. 
When words are masked, it will predict them based on the context words, entities and relations. 
For masked aligned entities, the goal is to predict them considering both linguistic and entity contexts, 
aligning the representation spaces of language and knowledge. 
In instances where relations between two distinct aligned entities are masked, 
the model is tasked with discerning the relationship between the two mentioned entities 
based on linguistic and entity contexts.}
However, the unified data structure above relies on heuristic approaches. To this end, some researchers propose a more general method.
\begin{figure}[htbp] 
	\centering 
	\includegraphics[width=3.5in]{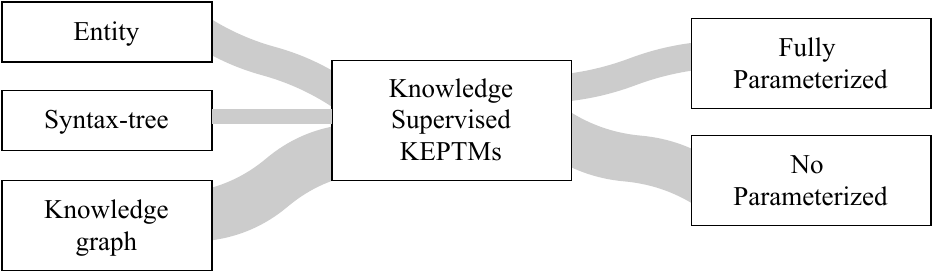} 
	\caption{The majority of knowledge-supervised KEPLMs use entities and knowledge graphs as supervisory signals, while a minority employ syntactic trees for the same purpose. Most of these models preserve symbolic knowledge in its original form, with only a few parameterizing it for integration into the model.}
	\label{section4}
\end{figure} 
Guan et al. \cite{116guan2020knowledge} and COMET \cite{new18bosselut2019comet} convert the relational triplets of KGs into meaningful sequences using specific templates. %
To generate reasonable stories with commonsense knowledge, Guan et al. transforms the commonsense triples in ConceptNet and ATOMIC into readable natural language sentences using a template-based method \cite{117levy2017zero}, and subsequently carries out post-training with these sentences by language model objectives \cite{116guan2020knowledge}. \par%
Notably, Daniel et al. find those entity representations generated by pre-trained language models exhibit strong generalization across inductive link prediction, entity classification, and information retrieval tasks \cite{new21daza2021inductive}. For instance, by transferring implicit knowledge from deep pre-trained language models, COMET learns to produce new objects coherent to its subject and relation and achieves automatic construction of commonsense knowledge bases. The reason lies in the learned representations capture both contextual information and knowledge. %
Even though data-structure-unified KEPLMs inject knowledge without extra engineering, they mainly focus on KGs and discard parts of KG structural information for the sake of compromise with the unified data structure. \par 

\subsection{Knowledge-supervised KEPLMs}
The characteristic of knowledge-supervised KEPLMs is that they choose keywords as training data under the supervision of KGs and learn its semantics by the power of the original PLMs, as shown in Fig. \ref{survey4}. 
\textcolor{black}{The correlation between knowledge injection methods, various types of knowledge, and the extent of knowledge parameterization is presented in Fig. \ref{section4}.}\par 

The supervision targets both entities and relational triplets. For instance, T5+SSM is pre-trained to reconstruct named entities and dates mined from Wikipedia by BERT and attain competitive results on open-domain question answering benchmarks \cite{new8roberts2020much}. Instead of utilizing independent entities, some models, like WKLM \cite{71xiong2019pretrained}, LIBERT \cite{75lauscher2020specializing}, and GLM \cite{78shen2020exploiting}, choose entities that exist in specific relations from  KGs as input data to guide models to capture it.
To directly derive real-world knowledge from unstructured text, WKLM designs the weakly supervised entity replacement detection training objective to force the model to learn the relation between entities. %
Compared to the MLM objective, the entity replacement task introduces stronger entity-level negative signals and preserves the linguistic correctness of the original sentence. %
Instead of using a single entity, LIEBRT takes entity pairs meeting semantic similarity constraints as training instances to enable BERT to understand the lexical-semantic relations \cite{75lauscher2020specializing}. %
Not limited to the specific relation, GLM drives pre-trained language models to capture implicit relations underlying raw text between related entities through the guidance of a KG \cite{78shen2020exploiting}.  %
However, entity representations generated by pre-trained language models exhibit weak generalization across linking prediction tasks over KGs. %
To this end, KEPLER  jointly optimizes parameters using both the knowledge and MLM objectives to obtain representations \cite{128wang2021kepler}. The key step involves KEPLER initializing knowledge embeddings using textual descriptions from RoBERTa rather than the KRL. %
Similar to KEPLER, K-ADAPTER also updates parameters by jointly learning knowledge and language information \cite{129wang2020k}. The distinction lies in K-ADAPTER's approach, as it designs adapters to store various infused knowledge types. This design keeps the original pre-trained model parameters fixed, effectively isolating interactions between different knowledge types and addressing the issue of catastrophic forgetting.. %

Up to this point, the aforementioned models have primarily focused on harnessing the encoder's capabilities to capture implicit relations among entities. However, for complex reasoning tasks, it becomes necessary to directly model relationships between entities. Entities can be easily annotated by incorporating Wikipedia hyperlinks and aligning them with entities in the knowledge graph, serving as conduits for knowledge injection. This, however, doesn't hold true for relationships due to the myriad ways they can be expressed. To directly model entity relationships, ERICA takes input sequences and concatenates them with knowledge graph relations  \cite{new13DBLP:conf/acl/QinLT00JHS020}. It  models entity relationships through discrimination pre-training tasks. Specifically, it employs two tasks: entity discrimination and relation discrimination. The former, given the head entity and relation, aims to infer the tail entity, while the latter distinguishes the semantic closeness of two relations. %
For syntactic parsing capabilities, LIMIT-BERT acquires language representations through linguistics supervised mask strategies \cite{new2zhou2020limit}. Given a sentence, it predicts its syntactic or semantic constituents using a pre-trained linguistic model, determining masking spans accordingly. To address the issue caused by the special token [mask], LIMIT-BERT employs encoder-based generators and discriminators similar to ELECTRA \cite{new3DBLP:conf/iclr/ClarkLLM20}. It uses tasks involving masked token prediction and replaced token detection to train the model. 
SKEP offers a unified sentiment representation for multiple sentiment analysis tasks \cite{new20tian2020skep}. Leveraging automatically-mined sentiment knowledge, SKEP embeds sentiment information at the word, polarity, and aspect levels into its representations through sentiment knowledge prediction objectives.  \par 

One major advantage of knowledge-supervised KEPLMs is their ease of implementation without requiring additional network architecture. Moreover, they offer flexibility in knowledge injection by choosing the target for prediction during pre-training or fine-tuning. For example, SKEP achieves promising results in various sentiment tasks by using sentiment words as masking targets, as sentiment analysis relies primarily on sentiment words and word polarity rather than entire texts. Another advantage of the knowledge injection approach is its ability to leverage contrastive learning techniques to enhance integration effectiveness. Contrastive learning, which has recently achieved state-of-the-art performance in NLP, enhances model robustness by distinguishing variance. Knowledge graphs can provide specific relationships, such as antonyms and synonyms, making them suitable as training data for contrastive learning. ERICA, for instance, captures in-text relational facts more effectively by utilizing entity and relation discrimination tasks. \par%

\subsection{Retrieval-based KEPLMs}
\begin{figure}[htbp]
	\centering
	\begin{minipage}[t]{0.5\linewidth}
		\centering
		\includegraphics[width=2.5in]{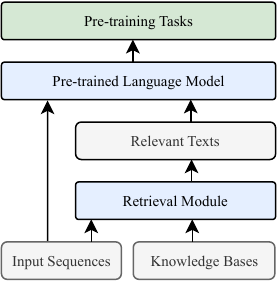}
		\caption{Infusing knowledge and PLMs via \\ retrieving relevant documents from external sources.} 
		\label{survey5}
	\end{minipage}%
	\begin{minipage}[t]{0.5\linewidth}
		\centering
		\includegraphics[width=2.5in]{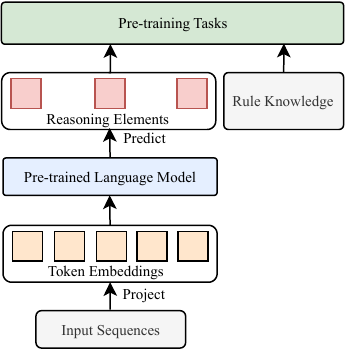}
		\caption{Combining rules to utilize knowledge through assembly.} 
		\label{survey6}
	\end{minipage}%
\end{figure}

\begin{figure}[htbp] 
	\centering 
	\includegraphics[width=3.5in]{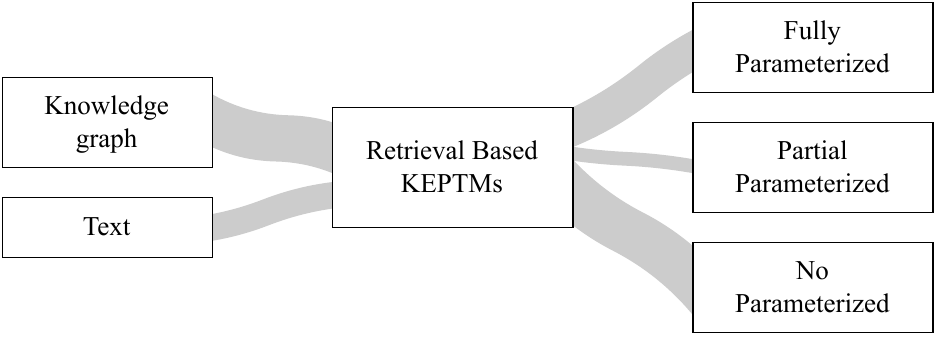} 
	\caption{The majority of retrieval-based KEPLMs use knowledge graphs as external reference, while a minority use texts. Most of these models either preserve symbolic knowledge in its original form or fully parameterize it, with only a small portion opting for partial parameterization.}
	\label{section5}
\end{figure}

\textcolor{black}{Given that retrieval-based KEPLMs are primarily designed for open-ended question answering tasks necessitating background knowledge related to questions, these models employ knowledge as an external reference. Contrary to KEPLMs that embed knowledge within model parameters, their primary objective is to retrieve, select, and encode the most relevant information from extensive knowledge repositories.} The process of knowledge integration is shown in Fig. \ref{survey5}.
This approach necessitates only a small training overhead.  Crucially, since there is no need to store large amounts of knowledge, such models allow for more efficient and convenient updating facing the frequent change of knowledge. 
\textcolor{black}{The correlation between knowledge injection methods, types of knowledge, and the extent of knowledge parameterization is presented in Fig. \ref{section5}.}\par 
For instance, it is more efficient to refer to critical information to judge rather than store all potentially related knowledge for question answering and generation tasks. %
KT-NET employs an attention mechanism to select desired knowledge from KGs adaptively and then fuses the selected knowledge to enable knowledge- and context-aware predictions for machine reading comprehension \cite{110yang2019enhancing}. %
It encodes the KG by the KRL \cite{41yang2014embedding} and learns to retrieve potentially relevant entities from WordNet and NELL \cite{111carlson2010toward} by fine-tuning. %
To supply factual knowledge, KGLM is constructed to render information from a local KG that is dynamically built by selecting and copying facts based on context from an external KG \cite{133logan2019barack}. \par %
The models we introduced above encode the KG with traditional methods of KRL that discard structure information. 
\textcolor{black}{Traditional KRL methods learn KG embeddings with solely first-order associations of entities and do not inherently encode nodes spanning varying distances.}
To solve it, various variants of GNN are employed to better capture global information of KGs. %
Lv et al. designed a graph-based model that extracts relational triplets from retrieved sentences and constructs self-defined graphs for them \cite{new19lv2020graph}. With customized graphs, the model adopts a graph convolutional network(GCN) to encode neighbor information into the node representations  and subsequently aggregates evidence with the graph attention mechanism for predicting the final answer. %
\textcolor{black}{The model starts by creating initial node representations with a PLM and updates them through the aggregation and combination mechanism of GCNs to incorporate neighbor information. 
Specifically, the model first generates node representations by averaging the hidden states of the entities with XLNet \cite{13yang2019xlnet}. 
Then the neighbor information is aggregated by taking the weighted average of the feature vectors of the immediate neighbors of a node, 
where each node's weight is determined by the attention mechanism. 
Once the neighbor information is aggregated, the GCN applies a neural network layer followed by a non-linear activation function to combine the aggregated information.} \par
In addition to their proficiency in question answering, Knowledge Graphs (KGs) also excel in generative tasks. To imbue GPT-2 with reasoning capabilities, GRF introduces ConceptNet as an external reference and generates conclusions based on the preceding context and the knowledge graph \cite{new14ji2020language}. At the heart of the model is the dynamic reasoning module, which computes the relevance between triplets and token embeddings to produce generated words. In the absence of context, Liu et al. proposed KG-BART, which generates coherent sentences using only a set of concepts \cite{new16liu2021kg}. It initially enhances token representations by considering a concept-reasoning graph structure. Subsequently, the model captures inherent correlations within and between concepts provided by a concept-expanding graph. The model excels in generating high-quality sentences, even when presented with unseen concept sets, by amalgamating knowledge graph and textual information.
As a supplement to structured knowledge, plain texts provide copious and comprehensive evidence. RAG generates responses by retrieving pertinent spans from external texts based on pre-trained seq2seq models \cite{new59DBLP:conf/nips/LewisPPPKGKLYR020}. Given a query, RAG utilizes the input sequence to retrieve the top K relevant texts and generates output conditioned on these latent documents, in conjunction with the input.  %
REALM enhances language models by retrieving and attending to documents in a modular and interpretable manner  \cite{142guu2020realm}. It comprises two vital components: a neural knowledge retriever implemented with the BERT framework, responsible for encoding input data and retrieving potentially useful documents, and a knowledge-augmented encoder implemented with a Transformer, used to incorporate entities from documents and predict words for question answering. 
TEK learns input text representations in conjunction with retrieved encyclopedic knowledge to capture and preserve factual knowledge linked to rare entities \cite{new9joshi2020contextualized}. Given the query and context, TEK retrieves a list of relevant sentences from multiple documents to enrich the input. Subsequently, the Transformer encoder operates on the input. To address the discrepancy between the input types seen during pre-training and inference, researchers conduct self-supervised pre-training on background knowledge-augmented input texts. \par 
Apart from their efficiency in knowledge utilization, retrieval-based models offer the advantage of interpretability in knowledge application. FaE establishes an explicit interface grounded in a neural language model to connect symbolically interpretable factual information with language representations, facilitating knowledge inspection and interpretation \cite{136verga2021adaptable}. 
Due to the separation of knowledge representations and language representations, FaE can modify the language model's output by altering only the non-parametric triplets, without requiring additional training. \par%
Despite their limited application scenarios, this approach allows for flexible encoding of knowledge as needed. More crucially, it enables the scrutiny of knowledge application when symbolic knowledge is preserved, thereby enhancing interpretability. For instance, FaE employs an external memory to store factual knowledge and observe corresponding predictions through the integration of different knowledge. %
Nevertheless, retrieval-based KEPLMs rely on labeled data for fine-tuning to acquire retrieval capabilities. Prompt-based models retrieve necessary knowledge from PLMs to tackle downstream tasks with a minimal number of samples. Despite their rapid development, these models may struggle to retrieve critical knowledge comprehensively. The inclusion of symbolic knowledge during prompt learning can alleviate the burden of parameter learning. For instance, PTR devises a prompt composed of embeddings and entities to achieve efficient learning  \cite{new60han2021ptr}. The fusion of prompt learning and knowledge enhances the synergy between few-shot learning and retrieval capabilities.

\subsection{Rule-guided KEPLMs}
As discussed above, presentation learning toward symbolic knowledge, like KGs, is a solution to bridge symbolic knowledge and pre-trained language models. By contrast, a prominent research direction is transforming the representation learned by PLMs into concepts and then reasoning with symbolic knowledge, holding effective learning of PLMs and symbolic knowledge's interpretability and accountability, as shown in Fig. \ref{survey6}.
Rule-guided KEPLMs focuses primarily on reasoning phase based on rules. Knowledge in these models is represented in symbolic form and integrated in a modular way. Their underlying characteristics allow the principled combination of robust learning and efficient reasoning, along
with interpretability offered by symbolic systems. 
\begin{figure}[htbp] 
	\centering 
	\includegraphics[width=3.5in]{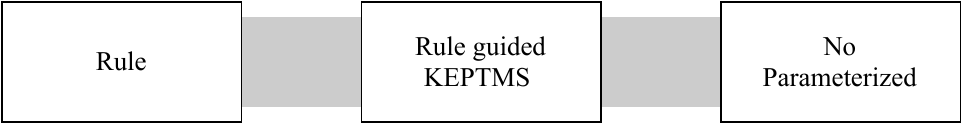} 
	\caption{Rule guide KEPLMs utilize rule knowledge by preserving the form of symbolic knowledge.}
	\label{section6}
\end{figure} 
\textcolor{black}{The correlation between knowledge injection methods,  types of knowledge, and the form of knowledge  is shown in Fig. \ref{section6}.}\par 

For instance, RuleBert strives to guide the PLM to reason on the deductive reasoning tasks by employing soft Horn rules \cite{saeed-etal-2021-rulebert}. Concretely, it predicts the correct probability of the reasoning output for the given soft rules and facts after fine-tuning. Compared to traditional language tasks, visual question answering requires reasoning and specific knowledge about the image subject and thus is a significantly more complex problem. To this end, Amizadeh et al. propose a more general mathematical formalism for visual question answering that is probabilistically derived from the first-order logic \cite{new32amizadeh2020neuro}. 
Prompt-tuning has been widely adopted on classification tasks \cite{new61liu2021pre}. However, manually designing language prompts is fallible, and those auto-generated are time-consuming to verify their effectiveness. To this end, PTR applies logic rules to construct prompts with several sub-prompts to make a trade-off \cite{new60han2021ptr}. It determines the subject and object entity types using the PLM and aggregates sub-prompts with logic rules to handle multi-class classification tasks.
\par %

Beyond first-order logic, some models \cite{new30wang2017explicit, new31wang2017fvqa} also utilize KGs to achieve reasoning. They relate concepts of a query image to the appropriate information in the KG to construct the local graph and reason subsequent correct answers over it. 
These approaches preserve the structure of the symbolic knowledge, thus achieving semantic inference and retrieval from a concept level. \par 

The main benefit of these models is their composability which entangles the representation and reasoning process, thereby enhancing the interpretability of the operational principles of KEPLMs.
\textcolor{black}{Although there are probing techniques to explore the interpretability of PLMs, they focus on what knowledge has been learned.} In contrast, this model type accomplishes the inspection and interpretation of knowledge usage by elucidating how a model could derive an answer.. %

We illustrate all introduced KEPLMs more details in Table \ref{overview1} and Table \ref{overview2}. %
\begin{table*}
	\centering
	\normalsize
	\caption{List of Representative KEPLMs}
	\label{overview1} 
	\begin{tabular}[c]{m{2.2cm}|m{2cm}m{2cm}m{2cm}m{2cm}m{2.2cm}} \hline 
		KEPLMs               & \begin{tabu}[c]{@{}l@{}}Pre-trained\\ Model\end{tabu} &  \begin{tabu}[c]{@{}l@{}}Granularity of\\ Knowledge\end{tabu} & KRL                                                                                                                                                 & \begin{tabu}[c]{@{}l@{}}Method of \\Injection  \end{tabu}        & \begin{tabu}[c]{@{}l@{}}Degree of \\Parameterization \end{tabu}  \\ \hline
		
		SenseBERT         & BERT                                                        & Entity    &           \textbackslash{}                                                                                                                                                     & \begin{tabu}[c]{@{}l@{}} Feature \\ Fused\end{tabu}                                                          & Fully       \\ \hline
		
		SentiLARE         & RoBERTa                                                     & Entity                & \textbackslash{}                                                     & \begin{tabu}[c]{@{}l@{}} Feature \\ Fused\end{tabu}                                                           & Fully       \\ \hline
		
		SKEP        & RoBERTa                                                        & Entity    &           \textbackslash{}                                                                                                                                                     & \begin{tabu}[c]{@{}l@{}} Knowledge\\ Supervised\end{tabu}                                                          & No       \\ \hline
		
		MWA        & BERT                                                        & Entity    &           \textbackslash{}                                                                                                                                                     & \begin{tabu}[c]{@{}l@{}} Knowledge \\ Supervised\end{tabu}                                                          & No       \\ \hline
		
		ZEN          & BERT                                                        & Entity          & Transformer                                                        &  \begin{tabu}[c]{@{}l@{}} Embedding \\ Combined\end{tabu}                                                           & Fully       \\ \hline

		T5+SSM              & T5                                                        & Entity         &      \textbackslash{}                                                       & \begin{tabu}[c]{@{}l@{}} Knowledge \\ Supervised\end{tabu}                                                           & Fully       \\ \hline
		
		ERNIE 1.0             & BERT                                                        & Entity         &      \textbackslash{}                                                       & \begin{tabu}[c]{@{}l@{}} Feature \\ Fused\end{tabu}                                                           & Fully       \\ \hline
		
		ERNIE 2.0             & BERT                                                        & Entity \&Text         &      \textbackslash{}                                                       & \begin{tabu}[c]{@{}l@{}} Feature \\ Fused\end{tabu}                                                           & Fully       \\ \hline
		
		LUKE         & BERT                                                     &   Entity                                                                                      & \textbackslash{}                                                                         & \begin{tabu}[c]{@{}l@{}}Knowledge \\ Supervised\end{tabu}          & No       \\ \hline \
		
		TEK          & RoBERTa                                                        & Text          & Transformer                                                       &  \begin{tabu}[c]{@{}l@{}} Retrieval \\ Based\end{tabu}                                                           & Fully       \\ \hline
		
		REALM             & BERT                                                        & Text             & \textbackslash{}                                                              & \begin{tabu}[c]{@{}l@{}} Retrieval \\ Based \end{tabu}                                                          & Fully      \\ \hline \
		
		RAG          & BART                                                        & Text          & \textbackslash{}                                                        &  \begin{tabu}[c]{@{}l@{}} Retrieval \\ Based\end{tabu}                                                           & Fully       \\ \hline
		
		Lv et al.     & XLNET                                                        & Text \& KG         & GNN                                                        &  \begin{tabu}[c]{@{}l@{}} Retrieval \\ Based\end{tabu}                                                           & Fully       \\ \hline\
		
		Syntax-BERT              & BERT                                                        & Syntax-tree         &      Transformer                                                      & \begin{tabu}[c]{@{}l@{}} Embedding \\ Combined\end{tabu}                                                           & Fully       \\ \hline\
		
		LIMIT-BERT              & BERT                                                        & Syntax-tree         &      \textbackslash{}                                                       & \begin{tabu}[c]{@{}l@{}} Knowledge \\ Supervised\end{tabu}                                                           & Fully       \\ \hline\
		
		Sachan et al.              & BERT                                                        & Syntax-tree         &      GNN                                                      & \begin{tabu}[c]{@{}l@{}} Embedding \\ Combined\end{tabu}                                                           & Fully       \\ \hline\
		
		WKLM              & BERT                                                        & KG             & \textbackslash{}                                                                                                                                                           & \begin{tabu}[c]{@{}l@{}}Knowledge \\ Supervised\end{tabu} & No       \\ \hline\
		
		LIBERT            & BERT                                                        & KG        & \textbackslash{}                                                                                                                                                     & \begin{tabu}[c]{@{}l@{}}Knowledge \\ Supervised \end{tabu}  & No       \\ \hline\
		
		GLM               & \begin{tabu}[c]{@{}l@{}}BERT or\\ RoBERTa\end{tabu}      & KG              & \textbackslash{}                                                                                                                 & \begin{tabu}[c]{@{}l@{}}Knowledge \\ Supervised \end{tabu}  & No       \\ \hline\
		
		COMET         & GPT2      & KG              & Transformer                                                                                                                 & \begin{tabu}[c]{@{}l@{}}Unified Data \\ Structure \end{tabu}  & Fully       \\ \hline
	\end{tabular}
	
\end{table*}

\begin{table*}
	\centering
	\normalsize
	\caption{List of Representative KEPLMs}
	\label{overview2} 
	\begin{tabular}{m{2.2cm}|m{2cm}m{2cm}m{2cm}m{2cm}m{2.2cm}} \hline
		KEPLMs               & \begin{tabu}[c]{@{}l@{}}Pre-trained\\ Model\end{tabu} &  \begin{tabu}[c]{@{}l@{}}Granularity of\\ Knowledge\end{tabu} & 
		KRL                                                                                                                                                 & \begin{tabu}[c]{@{}l@{}}Method of \\Injection  \end{tabu}        & \begin{tabu}[c]{@{}l@{}}Degree of \\Parameterization \end{tabu}  \\ \hline
		GRF               & LSTM      & KG              & GNN                                                                                                                & \begin{tabu}[c]{@{}l@{}}Retrieval \\ Based \end{tabu}  & Fully       \\ \hline\
		
		KG-BART           & BART                                                        &KG    & GAT                       & \begin{tabu}[c]{@{}l@{}} Retrieval \\ Based\end{tabu}                                                           & Fully       \\ \hline \

		KT-NET            & BERT                                                        & KG                                                              & BILINEAR                                                                  & \begin{tabu}[c]{@{}l@{}} Retrieval \\ Based \end{tabu}                                                           & Fully       \\ \hline
		KGLM              & LSTM                                                        & KG                                                              & TransE                                                        & \begin{tabu}[c]{@{}l@{}} Retrieval \\ Based \end{tabu}                                                          & Fully      \\ \hline \
		ERNIE             & BERT                                           &              KG         & TransE                                & \begin{tabu}[c]{@{}l@{}} Embedding \\ Combined\end{tabu}                                                           & Fully       \\ \hline
		
		KnowBERT          & BERT                                        & KG         & TuckER                                                                                                                            & \begin{tabu}[c]{@{}l@{}} Embedding \\ Combined\end{tabu}                                                           & Fully       \\ \hline\

		BERT-MK           & ERNIE                                                        &KG    & Transformer                       & \begin{tabu}[c]{@{}l@{}} Embedding \\ Combined\end{tabu}                                                           & Fully       \\ \hline \
		
		K-BERT            & BERT                                                        & KG                                                                                      & Transformer                                                                        & \begin{tabu}[c]{@{}l@{}}Unified Data \\ Structure\end{tabu}          & Fully       \\ \hline \
		
		CoLAKE            & RoBERTa                                                     & KG                                                                                      & Transformer                                                                        & \begin{tabu}[c]{@{}l@{}}Unified Data \\ Structure\end{tabu}          & Fully       \\ \hline \
		
		Guan et al. & GPT-2                                                       & KG     & Transformer                                                      & \begin{tabu}[c]{@{}l@{}}Unified Data \\ Structure\end{tabu}          & Fully       \\ \hline \
		
		KEPLER            & RoBERTa                                                  &   KG                                                                                      & \textbackslash{}                                                                         & \begin{tabu}[c]{@{}l@{}}Knowledge \\ Supervised\end{tabu}          & No       \\ \hline \
		
		K-ADAPTER         & RoBERTa                                                     &   KG                                                                                      & \textbackslash{}                                                                         & \begin{tabu}[c]{@{}l@{}}Knowledge \\ Supervised\end{tabu}          & No       \\ \hline \

		ERICA         & \begin{tabu}[c]{@{}l@{}}BERT or\\ RoBERTa\end{tabu}                                                       &   KG                                                                                      & \textbackslash{}                                                                         & \begin{tabu}[c]{@{}l@{}}Knowledge \\ Supervised\end{tabu}          & No       \\ \hline\
		
		ERNIE-ViL         & \begin{tabu}[c]{@{}l@{}}Transformer\end{tabu}                                                       &   KG                                                                                      & \textbackslash{}                                                                         & \begin{tabu}[c]{@{}l@{}}Knowledge \\ Supervised\end{tabu}          & No       \\ \hline\
		
		FaE               & BERT                                                        & KG          & \textbackslash{}                                                              & \begin{tabu}[c]{@{}l@{}} Retrieval \\ Based \end{tabu}                                                          & Partial      \\ \hline\
		
		PTR             &\begin{tabu}[c]{@{}l@{}}RoBERTa\end{tabu}                                                          & Rule             & \textbackslash{}                                                              & \begin{tabu}[c]{@{}l@{}} Rule \\ Guided \end{tabu}                                                          & No      \\ \hline
	\end{tabular}
\end{table*}

\section{Conclusion and Future Directions}
We analyzed and compared existing KEPLMs based on three criteria: granularity of knowledge, method of knowledge injection, and degree of knowledge parameterization. We then provided a detailed discussion on the second criterion. \par%

Most KEPLMs blend knowledge during pre-training, while a few do this during fine-tuning. However, compared to fine-tuning, the cost of integration during pre-training is much higher. Besides, choosing a consistent pre-training paradigm with PLMs can alleviate integration difficulties. For example, by masking out words that contain certain types of knowledge during generative pre-training. 

Feature-fused KEPLMs utilize entity information without introducing additional network and computational overhead.  They are simple to implement and suitable for tasks requiring fine-grained entity features. %
Despite more efforts, embedding-combined KEPLMs can store both entity and relation information and generalize knowledge-driven tasks like entity classification, relationship extraction, and knowledge completion. %
Knowledge-supervised KEPLMs facilitate knowledge infusion with minimal effort by selecting appropriate training data for the model's pre-training task. %
Retrieval-based and rule-guided KEPLMs help us understand how PLMs harness knowledge for downstream tasks, providing guidance for their optimal use and enhancement. %

Although KEPLMs have demonstrated their efficacy across various NLP tasks, they still face challenges stemming from the complexity of knowledge and language. %
\textcolor{black}{The diversity of knowledge types introduces varying representation approaches, making it challenging to achieve a unified knowledge representation and integration method. Moreover, the dynamic progression of knowledge, particularly when it exists in the form of model parameters, poses challenges for updating and maintaining its consistency and validity. Concurrently, the variability of language and morphological nuances complicate knowledge extraction and semantic alignment. For instance, in Chinese biomedical contexts, entity names often combine Chinese characters, English letters, and numbers, yielding multiple derivations through partial character omission. Consequently, this complicates the identification of entities and their potential relations. Moreover, words with the same suffixes in some fields might convey entirely different meanings, adding another layer of complexity.}
To overcome the aforementioned challenges, we suggest the following future directions. \par%

(1) Most of the KEPLMs we introduce focus on injecting factual or conceptual knowledge. 
There are other types of knowledge worthy of being considered. 
For instance, procedural and metacognitive knowledge also play a significant role in reasoning and judgment in the open world. 
Thus, a more attractive direction is to explore the utilization of the two types of knowledge above. 

(2) Grounded in semantic network representation, relational triplets have emerged as the most favored method for organizing knowledge. However, as we have discussed, more work has to be done for heterogeneous infusions caused by different representation methods of original training data and external knowledge. 
Besides semantic network representation, there are a multitude of knowledge representation methods presenting properties of knowledge in different forms. 
Therefore, searching a more general knowledge representation for different knowledge is promising. 

(3) Although retrieval-based and rule-guided KEPLMs make the procedure of decision-making transparent, 
they are designed for specific applications. 
Designing a more general KEPLM without destroying the inspection of symbolic knowledge 
will significantly improve interpretability. 

(4) The storage and updating of knowledge is barely considered by existing KEPLMs. 
In an environment where knowledge is changing rapidly, it is practical to store knowledge in less space and update it efficiently. 
The Adapter-based approach sets valuable examples for us. 
Designing a method that utilizes knowledge in a plug-and-play manner is essential. 

(5) Knowledge from different sources might present overlapping or even conflicting information. 
Ensuring that the knowledge from one source aligns with and complements the knowledge from another is crucial for the model's overall coherence and effectiveness. To address the alignment and consistency challenges, future models could incorporate mechanisms to verify the consistency of information across sources. This could involve cross-referencing and validating information from one source using another, ensuring that the integrated knowledge is coherent and consistent.

(6) Knowledge is usually extracted with multi-step processing. 
However, errors will propagate during this process, potentially diminishing model performance. 
Therefore, integrating knowledge excavated from the raw data to avoid information loss is a valuable direction.

(7) Despite their robust performance on entailment tasks, PLMs struggle with abductive reasoning \cite{new17bhagavatula2020abductive}. 
The previous work mainly focuses on formal logic that is too rigid to generalize for complex natural language. 
Integrating formal logic with PLMs offers a promising avenue for future research.

\begin{acks}
This research was supported by grants from the National Key Laboratory for Complex Systems Simulation Foundation (No.6142006200406).
\end{acks}

\bibliographystyle{ACM-Reference-Format}
\bibliography{refs}

\appendix

\end{document}